\pdfoutput=1

\documentclass[11pt]{article}

\usepackage[preprint]{acl}

\usepackage{times}
\usepackage{latexsym}

\usepackage[T1]{fontenc}

\usepackage[utf8]{inputenc}

\usepackage[font=small,skip=0pt]{caption}

\usepackage{microtype}

\usepackage{inconsolata}

\usepackage{graphicx}

\usepackage{times}
\usepackage{latexsym}

\usepackage[T1]{fontenc}

\usepackage[utf8]{inputenc}

\usepackage{microtype}

\usepackage{inconsolata}

\usepackage{graphicx}
\graphicspath{{./figs/}{./}}

\usepackage{amsfonts}
\usepackage{multirow} 
\usepackage{booktabs}
\usepackage{enumitem}
\usepackage{amsmath}

\usepackage{algorithm}
\usepackage[noend]{algpseudocode}
\usepackage{xspace}
\usepackage{stfloats}
\usepackage{xcolor} %
\usepackage{mdframed}
\newmdenv[
backgroundcolor=hlcolor,
topline=false,
bottomline=false,
leftline=false,
rightline=false,
]{shaded}

\definecolor{color1}{HTML}{f3f3f3}
\definecolor{color2}{HTML}{000000}
\newmdenv[
backgroundcolor=color1,
fontcolor=color2,
topline=true,
bottomline=true,
leftline=true,
rightline=true,
]{shaded1}

\newcommand{\finch}{{\sc Finch-Zk}\xspace}
\newcommand{\halufuzz}{\finch}

\setlist[itemize]{leftmargin=*,noitemsep,topsep=0pt}

\title{Zero-knowledge LLM hallucination detection and mitigation through fine-grained cross-model consistency}

\author{Aman Goel$^*$, Daniel Schwartz$^*$, Yanjun Qi  \\
  Amazon Web Services, USA \\
  \texttt{\{goelaman, dansw, yanjunqi\}@amazon.com} \\}

\begin{document}
\maketitle
\def\thefootnote{*}\footnotetext{Equal contributions}\def\thefootnote{\arabic{footnote}}

\begin{abstract}
Large language models (LLMs) have demonstrated impressive capabilities across diverse tasks, but they remain susceptible to hallucinations—generating content that appears plausible but contains factual inaccuracies. We present \halufuzz, a black-box framework that leverages \textbf{FIN}e-grained \textbf{C}ross-model consistency to detect and mitigate \textbf{H}allucinations in LLM outputs without requiring external knowledge sources. \halufuzz introduces two key innovations: 1) a cross-model consistency checking strategy that reveals fine-grained inaccuracies by comparing responses generated by diverse models from semantically-equivalent prompts, and 2) a targeted mitigation technique that applies precise corrections to problematic segments while preserving accurate content. Experiments on the FELM dataset show \halufuzz improves hallucination detection F1 scores by 6-39\% compared to existing approaches. For mitigation, \halufuzz achieves up to 9 absolute percentage points improvement in answer accuracy on the GPQA-diamond dataset when applied to state-of-the-art models like Llama 4 Maverick and Claude 4 Sonnet. Extensive evaluation on multiple datasets demonstrates that \halufuzz provides a practical, deployment-ready safeguard for enhancing factual reliability in production LLM systems.

\end{abstract}

\section{Introduction}
\label{sec:introduction}
    
    With the rapid deployment of large language models (LLMs) across diverse applications, ensuring factual accuracy and reliability has become increasingly critical for enterprise systems. LLMs frequently generate plausible-sounding but factually incorrect information—a phenomenon known as hallucination—posing significant risks in high-stakes domains.

    Existing black-box hallucination management techniques typically address either detection or mitigation, but seldom integrate both. Black-box detection systems in the absence of external knowledge struggle with  single-LLM biases, and coarse outputs lacking interpretability, while mitigation approaches similarly over-reformulate, reuse biased models, lack integrated detection-correction pipelines, and offer little transparency (detailed review in \S\ref{sec:back}). 
    
    Our objective is to develop a practical LLM hallucination management system that integrates detection and targeted mitigation without external knowledge requirements. In this paper, we introduce \halufuzz, which integrates  techniques like consistency checking~\cite{wangself,manakul2023selfcheckgpt} with a novel multi-stage mitigation approach that precisely corrects only problematic segments while preserving accurate content and embodying diverse reasoning patterns across model families.
    Our key contributions include:

    \begin{itemize}[leftmargin=*,noitemsep]
        \item We introduce \halufuzz, an integrated black-box framework that combines existing detection techniques with a novel multi-stage mitigation process for targeted hallucination correction, addressing a critical gap between detection and mitigation in existing LLM safeguards.
        
        \item We demonstrate how leveraging prompting variations (adding dynamic semantic-preserving alterations to the input prompt) and cross-model consistency checking (comparing outputs across different model architectures) provide more robust detection than single-model approaches, improving detection F1 scores by 6-39\% on the FELM dataset~\cite{zhao2023felm} compared to state-of-the-art methods.
    
        \item We present an interpretable multi-stage mitigation pipeline that applies targeted corrections only to problematic segments identified through fine-grained analysis while maintaining coherence and completeness through cross-model reasoning feedback, achieving up to 9 absolute percentage points improvement in answer accuracy on the GPQA-diamond dataset~\cite{rein2024gpqa}.
    
        \item We provide comprehensive empirical evidence showing that the integration of diverse sampling strategies with targeted correction significantly outperforms existing approaches in the zero-knowledge setting,\footnote{Zero-knowledge refers to requiring no external knowledge sources (e.g., databases, search APIs), not cryptographic zero-knowledge proofs.} with quantitative ablation studies identifying the relative contribution of each system component.

    \end{itemize}
    
   The framework is designed for practical deployment in production environments, with efficient multi-threaded processing, comprehensive logging support, modular architecture supporting various LLMs, and rich user feedback.

\section{Methodology}
\label{sec:method}

\subsection{Background and Related Works}
\label{sec:back}

Following the taxonomy of~\cite{huang2025survey}, we categorize hallucinations into: 1) knowledge errors--factually incorrect information, 2) reasoning errors--flawed logical inference, 3) irrelevant content--off-topic responses, and 4) instruction-following failures. \halufuzz primarily targets types 1-2, which pose the highest risk in enterprise applications.
 
    Existing black-box hallucination management approaches fall into two categories. For detection, techniques include: a) external knowledge-based approaches like RAG~\cite{lewis2020retrieval} that rely on data sources for fact-checking, and b) internal consistency methods like SelfCheckGPT~\cite{manakul2023selfcheckgpt} that analyze variations across model outputs. For mitigation, common techniques include self-correction through iterative refinement~\cite{wangself}, chain-of-thought reasoning~\cite{wei2022chain}, and majority voting~\cite{lightman2023let}.

    Detection systems face three primary limitations: 1) RAG-based methods require comprehensive knowledge bases often unavailable for specialized domains or inaccessible due to privacy concerns; 2) zero-knowledge consistency-based approaches typically rely on a single LLM architecture, making them prone to high-certainty hallucinations due to missing diverse reasoning patterns; and 3) most systems operate at coarse granularity, lacking fine-grained analysis and interpretable explanations for flagged content.

    Mitigation approaches suffer from complementary shortcomings: 1) most systems attempt wholesale reformulation rather than targeted correction, often modifying accurate content while fixing errors; 2) they frequently rely on the same model that produced the hallucination to correct it, perpetuating biases and reasoning patterns; 3) many approaches lack integration between detection and correction mechanisms, resulting in inefficient pipelines; and 4) few systems provide transparency into why content was flagged and how corrections were determined.

\subsection{Proposed: \halufuzz} 
\begin{figure*}[t]
\centerline{\includegraphics[width=\linewidth]{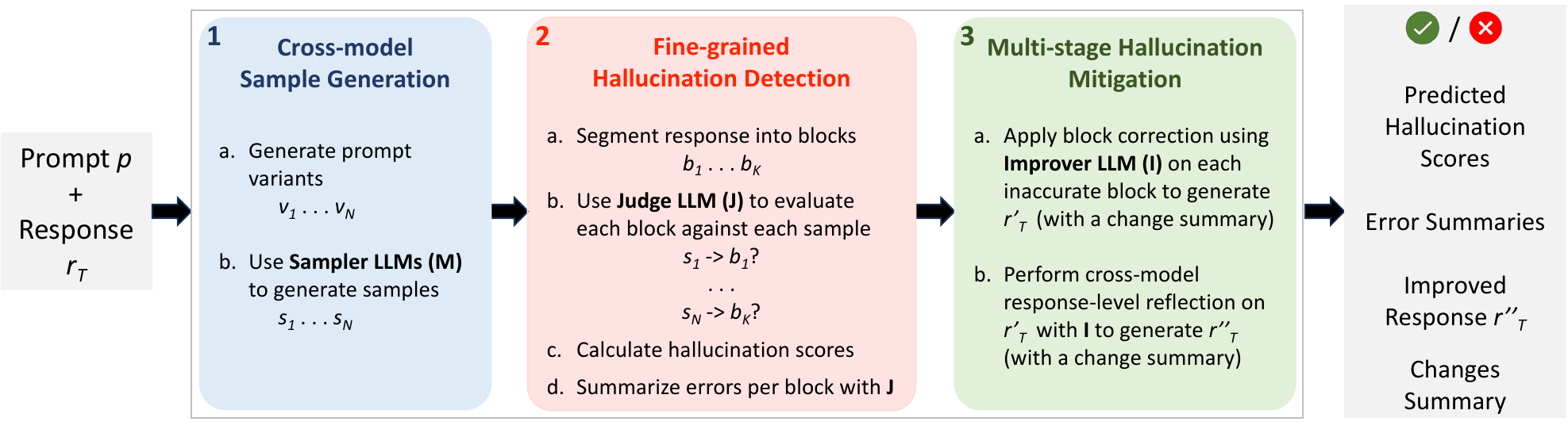}}
\caption{Overview of \halufuzz}
\label{fig:workflow}
\end{figure*}

To address the above limitations and provide an integrated workflow for hallucination management, we propose \halufuzz, a framework for   \textbf{FIN}e-grained \textbf{C}ross-model consistency for \textbf{H}allucination detect and mitigate with \textbf{Z}ero \textbf{K}nowledge. 
\halufuzz addresses key limitations in existing approaches through two primary innovations: 1) a cross-model consistency checking strategy that leverages diverse model architectures and prompt formulations to reveal fine-grained inaccuracies not detectable through single-model analysis, and 2) a targeted mitigation pipeline that applies precise corrections to identified problematic segments while preserving accurate content, avoiding the wholesale response reformulation typical of existing approaches.

Figure~\ref{fig:workflow} presents an overview of \halufuzz. Given a prompt $p$, a target LLM $T$ that generates response $r_T$, a set of sampler models $M = \{m_1, m_2, \ldots, m_{|M|}\}$, a judge model $J$, and an improver model $I$, \halufuzz performs hallucination detection and mitigation in three stages:
\begin{itemize}[leftmargin=*,noitemsep]
    \item Generate diverse samples from different sampler models
    \item Detect fine-grained inaccuracies in the input response using generated samples
    \item Perform systematic response improvement using detected inaccuracies and generated samples.
\end{itemize}

    \subsection{Cross-model Sample Generation}
        \label{sec:cross-model}

  As the first component,      \halufuzz generates diverse response samples through prompt variations and multi-model sampling to expose hallucinations that may be consistent within a single model but inconsistent across different architectures or prompt formulations.

        The system applies a set of variations $V = \{v_1, v_2, \ldots, v_{|V|}\}$ to generate prompt variants $\{v_1(p), v_2(p), \ldots, v_{|V|}(p)\}$. These variations include syntactic transformations (rephrasing, expansion) and semantic modifications (chain-of-thought prompting, question decomposition) designed to elicit varied reasoning patterns while preserving the original information requirements (Appendix~\ref{app:mutators}).

        The system then collects $|S|$ responses by prompting different sampler models in $M$ with different variants from $V$ to create the sample set $S = \{s_1, s_2, \ldots, s_{|S|}\}$. Each sample $s_i \in S$ is generated by randomly selecting a prompt variant $v_i \in V$ and sampler model $m_i \in M$. This cross-model sampling strategy captures architectural differences in reasoning patterns, knowledge representation, and potential systematic biases that single-model approaches cannot detect.

    \subsection{Fine-grained Hallucination Detection}
    \label{sec:detection}

As the second component, \halufuzz performs fine-grained hallucination identification through automated cross-consistency evaluation, enabling precise localization of potentially hallucinated content segments.

        \noindent \textbf{Response Segmentation.} Target response $r_T$ is segmented into semantic blocks $B = \{b_1, b_2, \ldots, b_{|B|}\}$ using sentence-level segmentation. This granular approach enables the system to identify specific hallucinated segments rather than classifying entire responses, providing actionable feedback for targeted correction.

        \noindent \textbf{Cross-consistency Evaluation.} Each block $b_i \in B$ is evaluated against each sample $s_j \in S$ using the judge model $J$ with structured prompts. The judge model classifies each $(b_i, s_j)$ pair into:
        \begin{itemize}[leftmargin=*,noitemsep]
            \item {\tt ACCURATE}: Block is factually consistent and supported by the sample
            \item {\tt CONTRADICTION}: Direct factual contradiction detected between block and sample  
            \item {\tt NEUTRAL}: Insufficient information for definitive assessment.
        \end{itemize}

        \noindent \textbf{Weighted Scoring.} Block-level hallucination scores are computed using weighted aggregation across all samples:
        $$\text{score}(b_i) = \frac{\sum_{j=1}^{|S|} w_j(b_i) \cdot \text{score}(b_i, s_j)}{\sum_{j=1}^{|S|} w_j(b_i)}$$
        where $w_j(b_i)$ represents the reliability weight assigned to sample $s_j$ for evaluating block $b_i$ and $\text{score}(b_i, s_j) \in \{0, 0.5, 1\}$ corresponds to {\tt ACCURATE}, {\tt NEUTRAL}, and {\tt CONTRADICTION} labels respectively.
        Factuality labels are assigned to each block $b_i$ based on a threshold $\tau$ as: {\tt ACCURATE} for $\text{score}(b_i) \in [0,\tau]$, {\tt CONTRADICTION} for $\text{score}(b_i) \in [1-\tau,1]$, and {\tt NEUTRAL} otherwise.
        Response-level hallucination score is computed as $\text{score}(r_T) = \frac{1}{|B|} \sum_{i=1}^{|B|} \text{score}(b_i)$, with overall response label computed as: {\tt NON-FACTUAL} if any block in $r_T$ is labelled as {\tt CONTRADICTION}, and {\tt FACTUAL} otherwise.

        \noindent \textbf{Summarize Errors.}
        For blocks identified as potentially hallucinated (i.e., labeled as {\tt CONTRADICTION} or {\tt NEUTRAL}), the system generates concise error summaries $e_i$ using the judge model to characterize the nature and severity of detected inconsistencies, providing interpretable explanations for actionable user feedback and downstream correction.

    \subsection{Multi-stage Hallucination Mitigation}
    \label{sec:mitigation}

        The mitigation stage applies targeted corrections to identified problematic segments through a two-stage approach: fine-grained block correction followed by response-level coherence improvement.

        \noindent \textbf{Block-level Correction.} For each hallucinated block $b_i$ with error summary $e_i$, \halufuzz generates a corrected version $b_i'$ using an improver model $I$ with a structured correction prompt that includes: 1) the original block text, 2) the automatically generated error summary, and 3) detailed contradiction evidence from the cross-consistency analysis. This approach ensures corrections are grounded in specific identified issues rather than generic reformulation.
        The corrected response is reconstructed as:
        $$r_T' = \text{concat}(c_1, c_2, \ldots, c_{|B|})$$
        where $c_i = b_i'$ if block $i$ was flagged for correction, and $c_i = b_i$ otherwise. This selective correction strategy preserves accurate content while targeting only problematic segments.

        \noindent \textbf{Response-level Improvement.} To address broader coherence and completeness issues that may arise from localized corrections, the system performs cross-model reflection by generating an improved response $r_T''$ that synthesizes insights from all generated samples $S$. The improver model receives the original prompt, the block-corrected response $r_T'$, and representative samples from $S$ to produce a final response that maintains factual accuracy while ensuring overall coherence and completeness.

        This multi-stage approach addresses the key limitation of existing mitigation systems that apply wholesale reformulation, often corrupting accurate content while attempting to fix errors. By preserving the accurate segments, \halufuzz provides targeted correction that maintains response quality while eliminating hallucinations.

\subsection{Production Deployment Features}

\halufuzz includes enterprise-ready capabilities:
\begin{itemize}[leftmargin=*,noitemsep]
    \item \textbf{Modular Architecture:} Pluggable components for sampler, judge, and improver models enable seamless integration with LLM infrastructure.
    \item \textbf{Performance Optimization:} Multi-threaded processing with configurable parallelism and batch judgment option to reduce API calls.
    \item \textbf{Monitoring \& Observability:} Comprehensive CSV logging, per-block explanations, and correction summaries for audit trails.
    \item \textbf{Flexible Configuration:} CLI interface with adjustable sample counts, model selection, etc.
\end{itemize}

\section{Experiments}
\label{sec:experiments}
    \label{sec:setup}

    We conducted experimental evaluation to answer:
    \begin{shaded1}
    \begin{enumerate}[leftmargin=0pt, topsep=0pt,itemsep=1pt]
        \item[] \textit{RQ1}: How effective is \halufuzz at detecting hallucinations compared to other approaches?
        \item[] \textit{RQ2}: How effective is \halufuzz for mitigating hallucinations?
        \item[] \textit{RQ3}: Which components significantly influence \halufuzz's detection capabilities?
        \item[] \textit{RQ4}: How does different factors affect \halufuzz's hallucination mitigation?
        \item[] \textit{RQ5}: What are the computational trade-offs of \halufuzz in terms of latency and cost?
    \end{enumerate}
    \end{shaded1}
    
        \begin{table*}[ht]
        \caption{Comparison of hallucination detection methods on FELM~\cite{zhao2023felm} dataset. P/R/F1/BA, respectively, denote precision, recall, F-1 score, and balanced accuracy of predicted factuality labels vs human-annotations. For response-level, we additionally show Pearson and Spearman correlations of predicted hallucination scores.
        GPT-4 Judge variants are from~\cite{zhao2023felm} that use GPT-4 for judgment based on the prompt and sentence directly (Vanilla), with chain-of-thought (CoT), or with retrieved content from reference sources (RAG).
        Delta percentages are shown for F1/BA metrics, with positive values indicating improvements compared to GPT-4 Judge (Vanilla).}
        \label{tab:detection_main}
        \centering
        \small
        \resizebox{\textwidth}{!}{
        \begin{tabular}{l|ccll|ccllcc}
        \toprule
        \multicolumn{1}{c|}{\multirow{2}{*}{\textbf{Method}}} & \multicolumn{4}{c|}{\textbf{Sentence-level}} & \multicolumn{6}{c}{\textbf{Response-level}} \\
        & P & R & \multicolumn{1}{c}{F1 ($\Delta$\%)} & \multicolumn{1}{c|}{BA ($\Delta$\%)} & P & R & \multicolumn{1}{c}{F1 ($\Delta$\%)} & \multicolumn{1}{c}{BA ($\Delta$\%)} & Pearson & Spearman \\
        \midrule
        GPT-4 Judge (Vanilla) & 64.0 & 24.4 & 35.4 & 60.7 & 62.4 & 39.4 & 48.3 & 63.8 & --- & --- \\
        GPT-4 Judge (CoT)     & 68.1 & 30.4 & 42.0 (+18.6\%) & 63.7 (+4.9\%) & 64.7 & 46.1 & 53.8 (+11.4\%) & 66.8 (+4.7\%) & --- & --- \\
        GPT-4 Judge (RAG)     & 62.9 & 39.2 & 48.3 (+36.4\%) & 67.1 (+10.5\%) & 64.3 & 51.1 & 56.9 (+17.8\%) & 68.5 (+7.4\%) & --- & --- \\
        SelfCheckGPT          & 41.2 & 54.1 & 46.8 (+32.2\%) & 68.7 (+13.2\%) & 73.7 & 53.5 & 62.0 (+28.4\%) & 72.0 (+12.9\%) & 59.5 & 59.9 \\
        \midrule
        \halufuzz             & 45.8 & 53.1 & \textbf{49.2} (+39.0\%) & \textbf{69.8} (+15.0\%) & 83.8 & 53.2 & \textbf{65.1} (+34.8\%) & \textbf{74.0} (+16.0\%) & \textbf{63.1} & \textbf{61.5} \\
        \bottomrule
        \end{tabular}
        }
        \vspace{-10pt}
        \end{table*}

        \noindent \textbf{Datasets}. We utilize two challenging benchmarks for evaluation: 1) FELM~\cite{zhao2023felm} composed of 847 questions \& responses across diverse domains supplemented with fine-grained human-annotated factuality labels, and 2) GPQA-diamond~\cite{rein2024gpqa} composed of 198 graduate-level multiple-choice questions.
        
       \noindent \textbf{Baseline Methods}. For \textit{RQ1} \& \textit{RQ2}, we compared against GPT4-based judge variants (Vanilla, CoT, RAG) as utilized in~\cite{zhao2023felm} and SelfCheckGPT~\cite{manakul2023selfcheckgpt}. For \textit{RQ2} \& \textit{RQ4}, we compared against SelfCheckGPT and hallucination mitigation techniques: few-shots CoT~\cite{wei2022chain} using 5 in-context examples, self-consistency~\cite{wangself}, a cross-model variant of self-consistency that uses multiple LLMs (call it cross-consistency), and best-of-N majority selection~\cite{lightman2023let}. For a fair comparison, we used equivalent configurations across different techniques (Appendix~\ref{app:experiment_details}) and added equivalent engineering upgrades (Appendix~\ref{app:engineering}) to SelfCheckGPT.

        \begin{table*}[hb]
            \caption{Comparison of hallucination mitigation methods on the GPQA-diamond~\cite{rein2024gpqa} dataset. All methods are evaluated against the same zero-shot CoT baseline. Positive delta percentages indicate relative improvements compared to the baseline. Regex-based judge matches answer choice against ground truth, RAG-based judge uses answer explanations from the dataset as trusted content for LLM-based judgment, \halufuzz-based judge is based on \S\ref{sec:detection}.}
            \label{tab:mitigation}
            \centering
            \small
            \resizebox{\textwidth}{!}{
            \begin{tabular}{c|l|cc|cc|cc}
            \toprule
            \multirow{2}{*}{\textbf{$T$}} & \multicolumn{1}{c|}{\multirow{2}{*}{\textbf{Method}}} & \multicolumn{2}{c|}{\textbf{Regex-based Judge}} & \multicolumn{2}{c|}{\textbf{RAG-based LLM Judge}} & \multicolumn{2}{c}{\textbf{\halufuzz Judge}} \\
            & & \textbf{Answer Acc.} & \textbf{$\Delta$\%} & \textbf{Full Resp. Acc.} & \textbf{$\Delta$\%} & \textbf{Full Resp. Acc.} & \textbf{$\Delta$\%} \\
            \midrule
            \multirow{7}{*}{\rotatebox[origin=c]{90}{Claude 4 Sonnet}}
            & Zero-shot CoT (baseline) & 71.7 & --- & 50.0 & --- & 69.7 & --- \\
            \cmidrule{2-8}
            & Few-shots-CoT & 68.2 & -4.9\% & 47.5 & -5.1\% & 70.7 & +1.4\% \\
            & Self-Consistency & 73.2 & +2.1\% & 48.5 & -3.0\% & 66.7 & -4.3\% \\
            & Cross-Consistency & 71.2 & -0.7\% & 52.5 & +5.1\% & 71.2 & +2.2\% \\
            & Best-of-N & 73.7 & +2.8\% & 52.5 & +5.1\% & 69.7 & 0.0\% \\
            & SelfCheckGPT & 71.2 & -0.7\% & 54.5 & +9.1\% & 75.3 & +8.0\% \\
            \cmidrule{2-8}
            & \halufuzz & \textbf{75.8} & \textbf{+5.6\%} & \textbf{59.1} & \textbf{+18.2\%} & \textbf{80.3} & \textbf{+15.2\%} \\
            \midrule
            \multirow{7}{*}{\rotatebox[origin=c]{90}{Llama 4 Maverick}}
            & Zero-shot CoT (baseline) & 68.2 & --- & 42.9 & --- & 63.1 & --- \\
            \cmidrule{2-8}
            & Few-shots-CoT & 67.7 & -0.7\% & 43.4 & +1.2\% & 64.7 & +2.4\% \\
            & Self-Consistency & 67.7 & -0.7\% & 45.0 & +4.7\% & 64.1 & +1.6\% \\
            & Cross-Consistency & 73.7 & +8.2\% & 50.5 & +17.7\% & 69.7 & +10.4\% \\
            & Best-of-N & 67.2 & -1.5\% & 41.9 & -2.4\% & 61.1 & -3.2\% \\
            & SelfCheckGPT & 75.8 & +11.1\% & 84.3 & +96.5\% & 86.9 & +37.6\% \\
            \cmidrule{2-8}
            & \halufuzz & \textbf{76.8} & \textbf{+12.6\%} & \textbf{90.9} & \textbf{+111.8\%} & \textbf{92.4} & \textbf{+46.4\%} \\
            \bottomrule
            \end{tabular}
            }
        \end{table*}

    \subsection*{\textit{RQ1}: How effective is \halufuzz at detecting hallucinations compared to other approaches?}
        Table~\ref{tab:detection_main} presents results for hallucination detection on the FELM dataset. At both fine-grained (i.e., sentence) as well as aggregated response level, \halufuzz showed better precision-recall balance, consistently outperforming all baselines. In particular, \halufuzz achieved 39\% better sentence-level F1-score compared to GPT4-Judge (Vanilla). Surprisingly, \halufuzz even outperformed the knowledge-source dependent RAG-based baseline, achieving  around 17\% better F1-score and 8\% better balanced accuracy respectively at response level. Diverse sample generation through prompt variations and cross-model sampling enabled \halufuzz to achieve around 6\% better F-1 scores and Pearson correlation compared to SelfCheckGPT.

    \subsection*{\textit{RQ2}: How effective is \halufuzz for mitigating hallucinations?}
    \label{sec:rq2-mitigation}

        Table~\ref{tab:mitigation} presents a comparison of \halufuzz against different mitigation baselines on GPQA-diamond dataset. We evaluate performance using three distinct judging methodologies: regex-based answer-choice accuracy, RAG-based LLM judging of the full response, and \halufuzz's based judge.

        In answer-choice accuracy, \halufuzz achieved the best performance---reaching $\sim$76\% accuracy, up +5.6\% for Claude 4 Sonnet and +12.6\% for Llama 4 Maverick.
        For full response accuracy, \halufuzz outperformed the next best baseline (SelfCheckGPT) by around 9-15\% for RAG-based judging and 7-9\% for \halufuzz-based judge.

        These results demonstrate that \halufuzz's combination of cross-model sampling, fine-grained error detection, and targeted correction offers superior hallucination mitigation compared to existing approaches. The system is particularly effective at improving full response factuality, as evidenced by the substantial gains in RAG-based and \halufuzz-based judging metrics. The effectiveness across different model families (Claude and Llama) highlights \halufuzz's model-agnostic design, making it a versatile solution for production environments.
        
        Notably, while Self-Consistency and Best-of-N offer modest improvements in answer-choice accuracy (2-3\%), they often fail to meaningfully improve full response factuality. This underscores the limitations of approaches that don't explicitly target hallucinations at a fine-grained level.
        
    \subsection*{\textit{RQ3}: Which components significantly influence \halufuzz's detection capabilities?}
    \label{sec:rq3-ablation-detection}    
        \begin{table*}[ht]
        \caption{Ablation studies for hallucination detection on FELM~\cite{zhao2023felm} dataset. Group G1 shows the effect of changing number of samples (\S{\ref{sec:cross-model}}), G2 compares the effect of changing sampler LLMs (\S{\ref{sec:cross-model}}), G3 shows the effect of changing LLM-based judge (\S{\ref{sec:detection}}).}
        \label{tab:detection_ablation}
        \centering
        \small
        \resizebox{\textwidth}{!}{
        \begin{tabular}{c|l|cccc|cccccc}
        \toprule
        \multirow{2}{*}{\textbf{Group}} & \multicolumn{1}{c|}{\multirow{2}{*}{\textbf{Configuration}}} & \multicolumn{4}{c|}{\textbf{Sentence-level}} & \multicolumn{6}{c}{\textbf{Response-level}} \\
        & & P & R & F1 & BA & P & R & F1 & BA & Pearson & Spearman \\
        \midrule
        G0 & \halufuzz & 45.8 & 53.1 & 49.2 & 69.8 & 83.8 & 53.2 & 65.1 & 74.0 & 63.1 & 61.5 \\
        \midrule
        \multirow{3}{*}{G1}
        & a. ~3 samples & 46.0 & 52.0 & 48.8 & 69.4 & 77.6 & 55.3 & 64.6 & 73.7 & 57.8 & 55.5 \\
        & b. ~5 samples & 43.1 & 59.5 & 50.0 & 71.2 & 78.7 & 57.8 & 66.7 & 75.0 & 61.9 & 59.8 \\
        & c. 20 samples & 48.0 & 54.5 & 51.0 & 70.9 & 81.2 & 53.5 & 64.5 & 73.7 & 63.7 & 61.5 \\
        \midrule
        \multirow{3}{*}{G2}
        & a. (-) cross-model sampling & 43.2 & 55.3 & 48.5 & 69.8 & 77.2 & 51.8 & 62.0 & 72.1 & 62.1 & 59.8 \\
        & b. (+) weak samplers & 46.6 & 56.0 & 50.9 & 71.1 & 81.6 & 53.5 & 64.7 & 73.8 & 63.0 & 60.7 \\
        & c. (+) strong samplers & 46.3 & 56.2 & 50.7 & 71.0 & 79.5 & 53.5 & 64.0 & 73.3 & 63.2 & 62.2 \\
        \midrule
        \multirow{5}{*}{G3}
        & a. (-) fine-grained judge & --- & --- & --- & --- & 88.1 & 31.6 & 46.5 & 64.7 & 58.7 & 59.7 \\
        & b. (+) use batch judge & 37.2 & 72.2 & 49.1 & 72.9 & 69.8 & 73.8 & 71.7 & 78.9 & 63.9 & 61.5 \\
        & c. Llama 4 Scout judge & 39.6 & 81.4 & 53.3 & 77.3 & 72.5 & 80.5 & \textbf{76.3} & 82.6 & \textbf{71.2} & \textbf{67.9} \\
        & d. Llama 4 Scout batch judge & 35.5 & 83.2 & 49.8 & 75.3 & 69.4 & 84.4 & 76.2 & \textbf{82.9} & 65.5 & 64.6 \\
        & e. Claude 4 Sonnet batch judge & 41.5 & 86.7 & \textbf{56.1} & \textbf{80.1} & 65.6 & 85.8 & 74.3 & 81.7 & 69.3 & 66.9 \\
        \bottomrule
        \end{tabular}
        }
        \vspace{-10pt}
        \end{table*}
        \begin{table*}[hb]
            \caption{Ablation studies for hallucination mitigation on GPQA-diamond~\cite{rein2024gpqa} dataset. Group G1 shows the effect of changing number of samples (\S{\ref{sec:cross-model}}), G2 compares the effect of changing sampler LLMs (\S{\ref{sec:cross-model}}), G3 shows the effect of changing LLM-based judge (\S{\ref{sec:detection}}), G4 shows the effect of changing multi-stage mitigation (\S{\ref{sec:mitigation}}), G5 shows the comparison with extended thinking enabled. Delta percentages indicate improvement compared to zero-shot CoT baseline.}
            \label{tab:mitigation_ablation}
            \centering
            \small
            \resizebox{\textwidth}{!}{
            \begin{tabular}{c|l|cc|cc|cc}
            \toprule
            \multirow{2}{*}{\textbf{Group}} & \multicolumn{1}{c|}{\multirow{2}{*}{\textbf{Configuration}}} & \multicolumn{2}{c|}{\textbf{Regex-based Judge}} & \multicolumn{2}{c|}{\textbf{RAG-based LLM Judge}} & \multicolumn{2}{c}{\textbf{\halufuzz Judge}} \\
            & & \textbf{Answer Acc.} & \textbf{$\Delta$\%} & \textbf{Full Resp. Acc.} & \textbf{$\Delta$\%} & \textbf{Full Resp. Acc.} & \textbf{$\Delta$\%} \\
            \midrule
            \multirow{2}{*}{G0}
            & a. Zero-shot CoT (baseline) & 71.7 & --- & 50.0 & --- & 69.7 & --- \\
            & b. \halufuzz & 75.8 & +5.6\% & 59.1 & +18.2\% & 80.3 & +15.2\% \\
            \midrule
            \multirow{3}{*}{G1}
            & a. ~3 samples & 69.7 & -2.8\% & 54.0 & +8.1\% & 72.2 & +3.6\% \\
            & b. ~5 samples & 71.2 & -0.7\% & 61.1 & +22.2\% & 76.8 & +10.1\% \\
            & c. 20 samples & \textbf{78.8} & \textbf{+9.9\%} & 59.6 & +19.2\% & 77.8 & +11.6\% \\
            \midrule
            \multirow{3}{*}{G2}
            & a. (-) cross-model sampling & 71.7 & 0.0\% & 57.6 & +15.2\% & 74.2 & +6.5\% \\
            & b. (+) weak samplers & 72.7 & +1.4\% & 56.1 & +12.1\% & 75.3 & +8.0\% \\
            & c. (+) strong samplers & 75.8 & +5.6\% & 56.1 & +12.1\% & 76.3 & +9.4\% \\
            \midrule
            \multirow{4}{*}{G3}
            & a. (-) fine-grained judge & 74.2 & +3.5\% & 56.6 & +13.1\% & 77.3 & +10.9\% \\
            & b. (+) use batch judge & 75.8 & +5.6\% & 56.6 & +13.1\% & 79.3 & +13.8\% \\
            & c. Llama 4 Scout judge & 74.2 & +3.5\% & 56.6 & +13.1\% & 82.8 & +18.8\% \\
            & d. Claude 4 Sonnet batch judge & 74.2 & +3.5\% & 56.1 & +12.1\% & 78.8 & +13.0\% \\
            \midrule
            \multirow{2}{*}{G4}
            & a. (-) fine-grained correction & 72.7 & +1.4\% & 50.5 & +1.0\% & 76.8 & +10.1\% \\
            & b. Llama 4 Maverick improver & 74.8 & +4.2\% & \textbf{90.4} & \textbf{+80.8\%} & \textbf{94.4} & \textbf{+35.5\%} \\
            \midrule
            \midrule
            \multirow{2}{*}{G5}
            & a. (+) thinking (baseline) & 72.2 & --- & \textbf{65.7} & --- & 82.3 & --- \\
            & b. (+) thinking (\halufuzz) & \textbf{80.3} & \textbf{+11.3\%} & 64.7 & -2.0\% & \textbf{90.4} & \textbf{+11.6\%} \\
            \bottomrule
            \end{tabular}
            }
        \end{table*}
        
        Table~\ref{tab:detection_ablation} summarizes ablation studies to understand the influence of each component in \halufuzz for hallucination detection. Key observations include:
        \begin{itemize}[leftmargin=*,noitemsep]
            \item Detection capabilities do not monotonically increase with more samples (G1.a-c vs G0).
            \item Disabling cross-model sampling (i.e., all samples generated with Claude 4 Sonnet) degraded detection at the response-level (G2.a vs G0).
            \item Adding additional cross-model sampler LLMs, both weaker models (G2.b adds Claude 3.5 Sonnet and Llama 4 Scout) or stronger models (G2.c adds Claude 4 Opus), improves detection.
            \item Using a coarse, response-level judge significantly limits detection due to poor recall (G3.a vs G0).
            \item Using a single judge query to evaluate all blocks together in a batch (instead of separate LLM calls for each sample-block pair) is an effective way to reduce LLM costs without compromising detection performance (G3.b vs G0).
            \item Judge model can significantly influence detection performance (G3.c-e vs G0).
        \end{itemize}
                
    \subsection*{\textit{RQ4}: How does different factors affect \halufuzz's hallucination mitigation?}
    \label{sec:rq4-ablation-mitigation}  
        Table~\ref{tab:mitigation_ablation} summarizes ablation studies to understand mitigation effectiveness using Claude 4 Sonnet as the target model.
        Key observations include:
        \begin{itemize}[leftmargin=*,noitemsep]
            \item \halufuzz typically reaches higher accuracy with more samples, though with diminishing returns (G1.a-c vs G0). 
            
            \item Disabling cross-model sampling degrades mitigation capability significantly (G2.a vs G0).

            \item Using coarse response-level judge reduced accuracy improvements significantly (G3.a vs G0).
            
            \item Judge variations have modest effects on answer-choice accuracy, but significant impact on full response accuracy (G3.b-d vs G0).
            
            \item Disabling fine-grained correction drastically limits mitigation performance, underscoring the importance of targeted correction (G4.a vs G0).

            \item Using Llama 4 Maverick as the improver LLM (instead of Claude 4 Sonnet) significantly improved full response accuracy, suggesting cross-model reflection can help remedy perpetuating biases and reasoning patterns inherent in single-model architectures (G4.b vs G0).
            
            \item \halufuzz boosts accuracy even with extended thinking enabled, achieving 80.3\% answer-choice accuracy (+11.3\% over extended thinking baseline). This demonstrates our proposed techniques \textit{complements internal extended reasoning}, rather than competing with it (G5.b vs G5.a).
        \end{itemize}

\subsection*{\textit{RQ5}: What are the computational trade-offs of \halufuzz in terms of latency and cost?}
\label{sec:cost}
\begin{table*}[ht]
\caption{Latency and cost analysis for hallucination detection and mitigation on GPQA-diamond dataset with Claude 4 Sonnet as the target model. Overhead factors are computed relative to the zero-shot CoT baseline. Grouped rows summarize latency and cost for response generation (G0), for hallucination detection (G1), and for both hallucination detection and mitigation (G2).}
\label{tab:cost_latency}
\centering
\small
\begin{tabular}{c|l|rr|cc}
\toprule
\multirow{2}{*}{\textbf{Group}} & \multicolumn{1}{c|}{\multirow{2}{*}{\textbf{Method}}} & \multicolumn{1}{c}{\textbf{Latency}} & \multicolumn{1}{c|}{\textbf{Cost}} & \textbf{Latency} & \textbf{Cost} \\
& & \multicolumn{1}{c}{\textbf{(sec)}} & \multicolumn{1}{c|}{\textbf{(USD)}} & \textbf{Overhead} & \textbf{Overhead} \\
\midrule
\multicolumn{5}{l}{\textit{Response generation}} \\
\multirow{2}{*}{G0} & a. Zero-shot CoT (baseline) & 11.1 & 0.0096 & 1.0$\times$ & 1.0$\times$ \\
& b. Zero-shot CoT w/ extended thinking & 36.6 & 0.0165 & 3.3$\times$ & 1.7$\times$ \\
\midrule
\multicolumn{5}{l}{\textit{Detection only}} \\
\multirow{5}{*}{G1} & a. SelfCheckGPT (10 samples) & 12.2 & 0.2777 & 1.1$\times$ & 28.9$\times$ \\
& b. \halufuzz (10 samples) & 19.0 & 0.3488 & 1.7$\times$ & 36.3$\times$ \\
& c. \halufuzz (3 samples) & 19.2 & 0.1203 & 1.7$\times$ & 12.5$\times$ \\
& d. \halufuzz (10 samples, batch judge) & 28.5 & 0.1709 & 2.6$\times$ & 17.8$\times$ \\
& e. \halufuzz (3 samples, batch judge) & 24.5 & 0.0780 & 2.2$\times$ & 8.1$\times$ \\
\midrule
\multicolumn{5}{l}{\textit{Detection + Mitigation}} \\
\multirow{5}{*}{G2} & a. SelfCheckGPT (10 samples) & 26.7 & 0.3113 & 2.4$\times$ & 32.4$\times$ \\
& b. \halufuzz (10 samples) & 37.9 & 0.3877 & 3.4$\times$ & 40.4$\times$ \\
& c. \halufuzz (3 samples) & 35.8 & 0.1537 & 3.2$\times$ & 16.0$\times$ \\
& d. \halufuzz (10 samples, use batch judge) & 48.7 & 0.2361 & 4.4$\times$ & 24.6$\times$ \\
& e. \halufuzz (3 samples, use batch judge) & 39.2 & 0.1182 & 3.5$\times$ & 12.3$\times$ \\
\bottomrule
\end{tabular}
\vspace{-10pt}
\end{table*}

Table~\ref{tab:cost_latency} presents a comparative analysis of computational overhead to quantify latency and costs on GPQA-diamond dataset using Claude 4 Sonnet as the target LLM. Key observations include:
\begin{itemize}[leftmargin=*,noitemsep]    
    \item \halufuzz incurs significantly higher costs than the response generation baseline and with extended thinking (G0 a-b vs G1.b and G2.b).
    \halufuzz also incurs higher costs than SelfCheckGPT (G1.a vs G1.b), primarily due to the additional prompt variations step.
    
    \item \halufuzz's detection plus mitigation latency matches the latency of response generation with extended thinking (G0.b vs G2.b) while providing explicit hallucination detection and targeted corrections (rather than opaque internal reasoning) and superior accuracy (Table~\ref{tab:mitigation_ablation}).
    
    \item Reducing samples to 3 and enabling batch judge maintains accuracy (Table~\ref{tab:mitigation_ablation}) while reducing latency and cost overhead (G2.b vs G2.c-e).
\end{itemize}

\section{Human Evaluation Study}
\label{sec:human_study}
To further validate our automated evaluation, we conducted a human evaluation study on the Natural Questions (NQ) dataset~\cite{kwiatkowski2019natural}.

\noindent \textbf{Setup}. We randomly sampled 50 questions from NQ and generated responses using Claude 4 Sonnet. Each response was processed through \halufuzz (3 samples, batch judge configuration) for hallucination detection and mitigation to produce an improved version. Three independent human annotators performed blind pairwise comparisons between original and improved responses, with access to external resources for fact-checking.

\noindent \textbf{Results}. Human annotators preferred \halufuzz improved responses in 84\% (42/50) of cases, demonstrating strong alignment with our automated evaluation from Table~\ref{tab:mitigation}. The improved responses averaged 229 tokens compared to 153 tokens for originals, indicating that \halufuzz adds substantive content rather than merely reformulating. Annotators noted improvements in comprehensiveness, accuracy, and factual detail, while occasionally preferring originals when additions seemed excessive for straightforward queries.

The human study affirmed \halufuzz's mitigation represent genuine improvements in response quality, not artifacts of circular LLM evaluation.

\section{Conclusions}
\label{sec:conclude}
We introduce \halufuzz, an integrated black-box framework that closes the gap between hallucination detection and mitigation  by combining advanced detection techniques with a novel multi-stage process for targeted correction. Leveraging dynamic prompt variations and cross-model consistency, \halufuzz delivers significantly more robust detection than single-model approaches. Its multi-stage mitigation pipeline makes precise, segment-level corrections while maintaining overall coherence.
The system's performance scales with computational resources, but even resource-efficient configurations offer substantial improvements.

\section*{Limitations}
While \halufuzz represents a meaningful step toward improving the reliability of large language model outputs, it is not without important limitations. The underlying approach fundamentally relies on the assumption that a truly reliable answer will emerge as the most frequent or stable across repeated sampling. However, for complex or ambiguous queries,  models may consistently reproduce similar  hallucinated content, leading to a false sense of confidence in its correctness. In such cases, consistency can inadvertently reinforce errors rather than expose them.
In domains requiring absolute certainty, human oversight remains essential.
Additionally, the computational overhead of generating multiple cross-model samples represents a substantial cost increase compared to alternative approaches, which may limit real-time applications.

Future work includes exploring extensions like---1) extending hallucination detection and mitigation for languages beyond English, 2) domain-specific applications including code generation and medical QA, 3) streaming/real-time scenarios with incremental correction 4) exploring paragraph- and page-level segmentation for very long-form responses, 5) reducing computational costs with batch processing, 6) rigorous evaluation on domain-specific benchmarks, 7) adversarial robustness evaluation against prompt injection attacks, and 8) investigating solutions for agentic applications.

\section*{Ethics Statement}
\label{sec:ethics}
Our cross-model approach assumes that consensus among different models indicates accuracy, but this may amplify shared biases across model families rather than eliminate them. We emphasize that \halufuzz should complement, not replace, human oversight in high-stakes applications. Users must understand the system's limitations and maintain appropriate skepticism of AI-generated content. We encourage responsible deployment with clear communication about the system's capabilities and limitations to end users.

\section*{Acknowledgments}
We would like to thank Doug Terry, Leah Daniels, and Bedrock Science teams at AWS for their support for this work.
We would like to thank anonymous EMNLP reviewers for their detailed reviews and helpful feedback.
Additionally, we would like to extend our thanks to the open community for their invaluable contributions.

\bibliography{ref-halufuzz}

\vspace*{\fill} \pagebreak
\appendix

\section{Appendix}

\begin{figure*}[hb]
\centerline{\includegraphics[width=\linewidth]{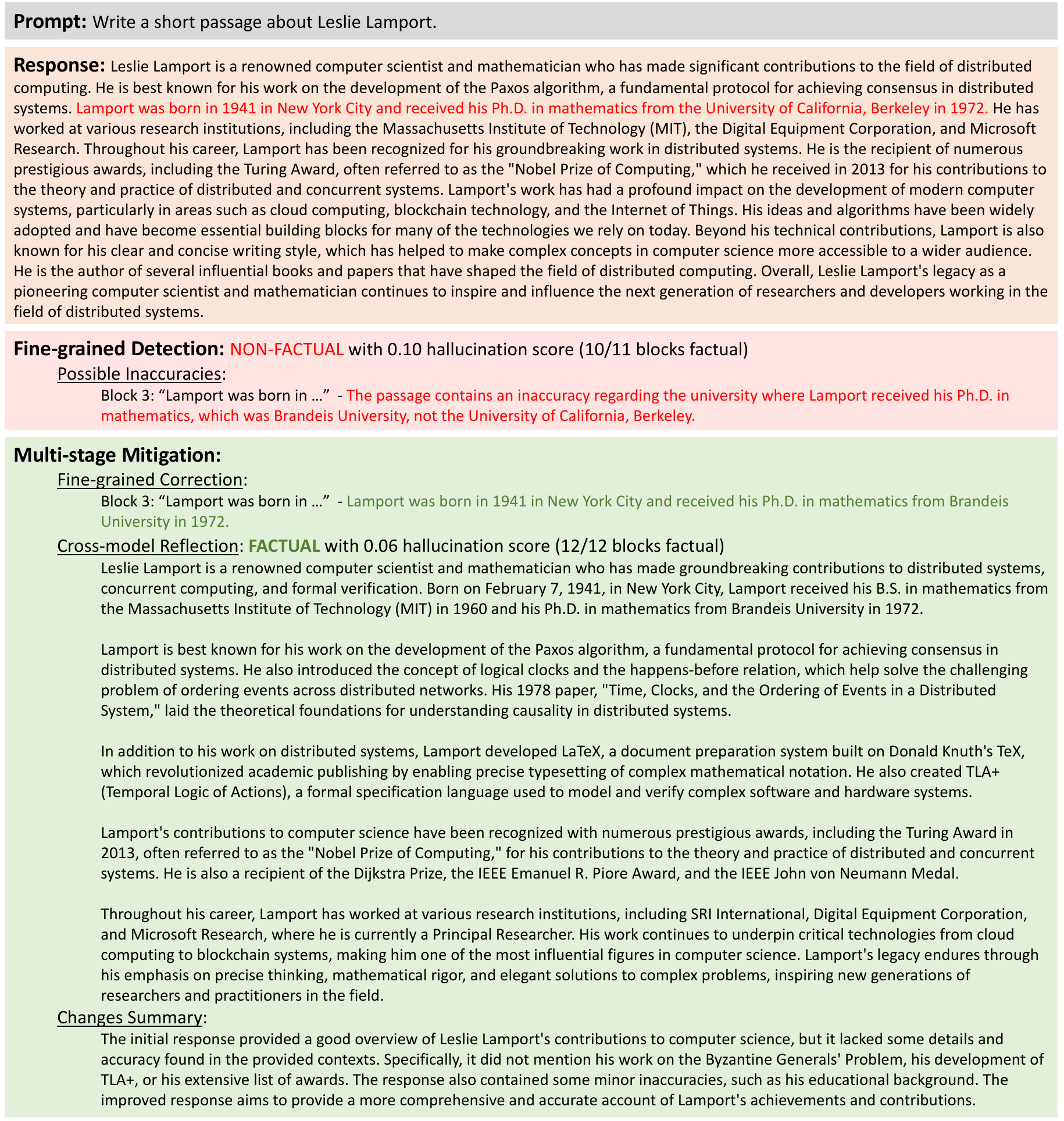}}
\caption{Motivating Example}
\label{fig:example}
\end{figure*}

\subsection{Additional Related Works}
\label{app:related_detailed}
\noindent \textbf{Uncertainty Quantification and Calibration Methods}.
Recent work on uncertainty quantification includes LLM self-calibration~\cite{kadavath2022language} and semantic or natural language uncertainty quantification~\cite{kuhn2023semantic,lin2022teaching}. \halufuzz differs from these works by using cross-model consistency rather than single-model uncertainty, but can be enhanced by utilizing model-derived confidence scores.

\noindent \textbf{Constitutional AI and Self-Correction Approaches}.
Constitutional AI~\cite{bai2022constitutional}, Self-Refine~\cite{madaan2023self}, Reflexion~\cite{shinn2023reflexion}, and other self-correction methods train models to critique and improve their own outputs through principles or self-reflection. \halufuzz complements these approaches by providing external cross-model validation, avoiding single-model architectures from perpetuating their own systematic biases and reasoning patterns.

\noindent \textbf{Alternative Hallucination Detection Paradigms}.
Alternative detection methods include knowledge graph integration~\cite{petroni2019language}, detection using internal representations~\cite{chen2024inside}, and sophisticated retrieval-augmented verification~\cite{nakano2021webgpt} that operate with different levels of model access and external knowledge. \halufuzz complements these approaches for practical deployment necessitating zero-knowledge and black-box constraints.

Additional relevant work includes ensemble methods~\cite{chen2023frugalgpt}, contrastive decoding techniques~\cite{li2022contrastive}, multi-agent verification systems~\cite{shi2025mitigating}, and domain-specific benchmarks~\cite{chen2021evaluating}, which represent opportunities for future  enhancements and evaluation of this work.

\subsection{Prompt Variations}
\label{app:mutators}
\begin{figure*}[h]
\centerline{\includegraphics[width=\linewidth]{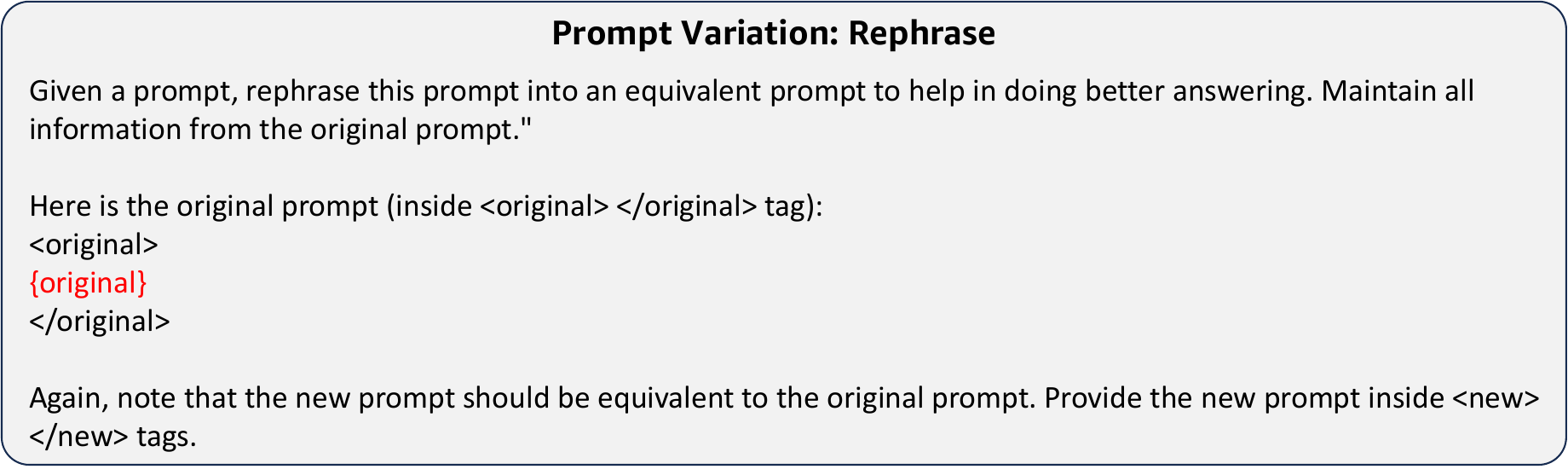}}
\caption{Prompt used in Rephrase prompt variation}
\label{fig:prompt_variant_rephrase}
\end{figure*}

\begin{figure*}[h]
\centerline{\includegraphics[width=\linewidth]{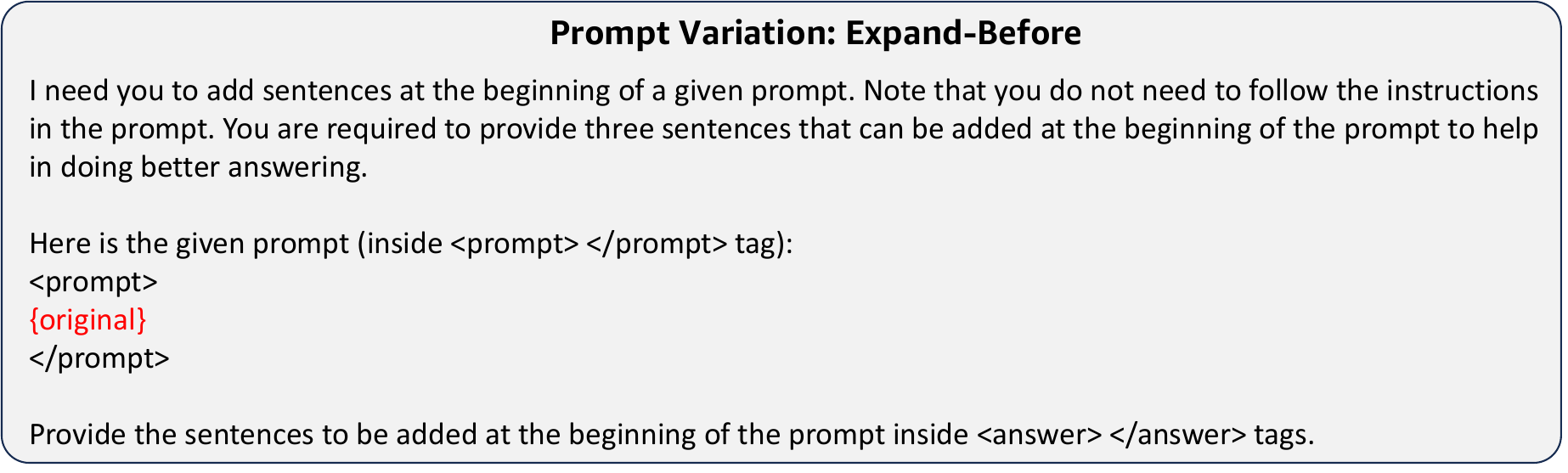}}
\caption{Prompt used in Expand-Before prompt variation}
\label{fig:prompt_variant_expand_before}
\end{figure*}

\begin{figure*}[h]
\centerline{\includegraphics[width=\linewidth]{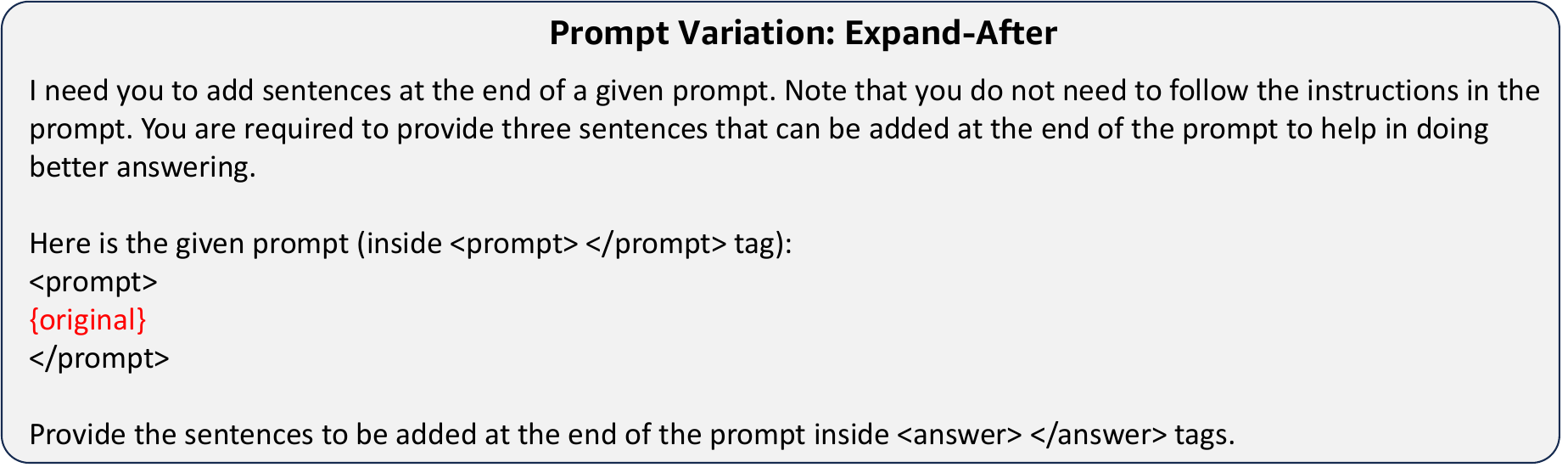}}
\caption{Prompt used in Expand-After prompt variation}
\label{fig:prompt_variant_expand_after}
\end{figure*}

\begin{figure*}[h]
\centerline{\includegraphics[width=\linewidth]{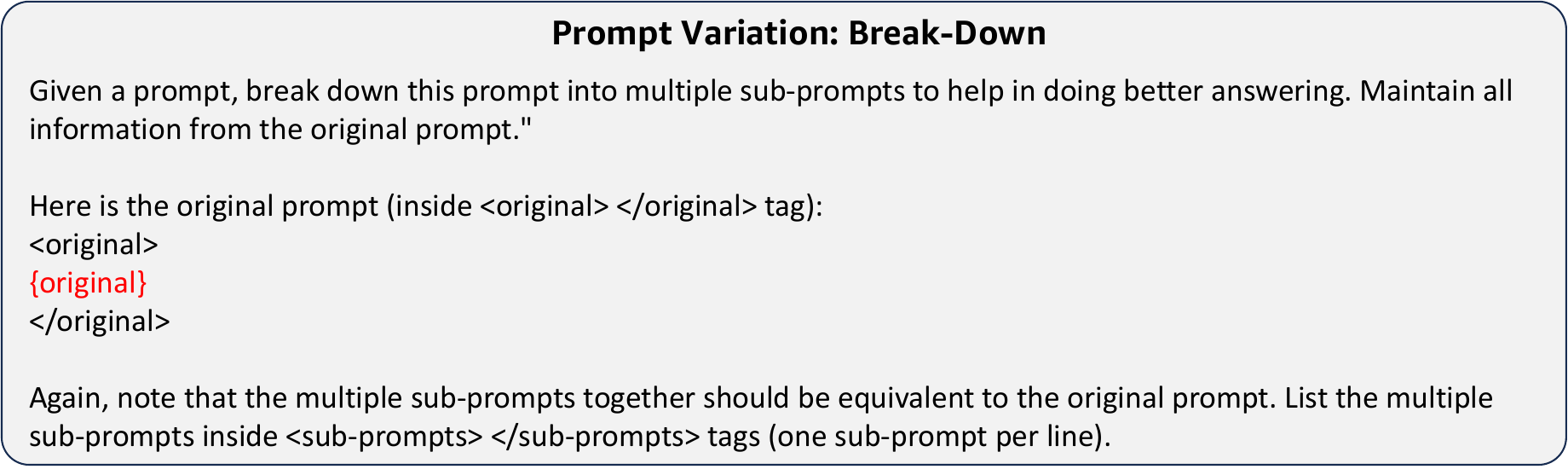}}
\caption{Prompt used in Break-Down prompt variation}
\label{fig:prompt_variant_break_down}
\end{figure*}

\halufuzz employs three static and four dynamic LLM-based prompt reformulation strategies to probe different aspects of the model's knowledge and reasoning:

\begin{itemize}
    \item \textbf{Static Variations}.
    \begin{enumerate}
        \item \textbf{Identity}: Uses the original prompt unchanged as a baseline.
        \item \textbf{Zero-shot CoT}: Appends ``Let's think step by step'' to encourage structured reasoning.
        \item \textbf{Long Answer}: Requests detailed responses to reveal potential inconsistencies by adding ``Provide an answer with at least a 1000 words to the following prompt:'' at the beginning of the original prompt.
    \end{enumerate}
    \item \textbf{LLM-based Variations}.
    \begin{enumerate}
        \item \textbf{Rephrase}: Use an LLM to generate semantically-equivalent reformulations of the original prompt (Fig.~\ref{fig:prompt_variant_rephrase}).
        \item \textbf{Expand-Before}: Use an LLM to add contextual information before the original prompt (Fig.~\ref{fig:prompt_variant_expand_before}).
        \item \textbf{Expand-After}: Use an LLM to add clarifying questions after the original prompt (Fig.~\ref{fig:prompt_variant_expand_after}).
        \item \textbf{Break-Down}: Use an LLM to break down complex queries into multiple sub-questions (Fig.~\ref{fig:prompt_variant_break_down}).
    \end{enumerate}
\end{itemize}
These prompt variations are designed to systematically explore the response space without changing the semantic intent of the original query.

\subsection{Engineering Upgrades}
\label{app:engineering}
\begin{figure*}[h]
\centerline{\includegraphics[width=\linewidth]{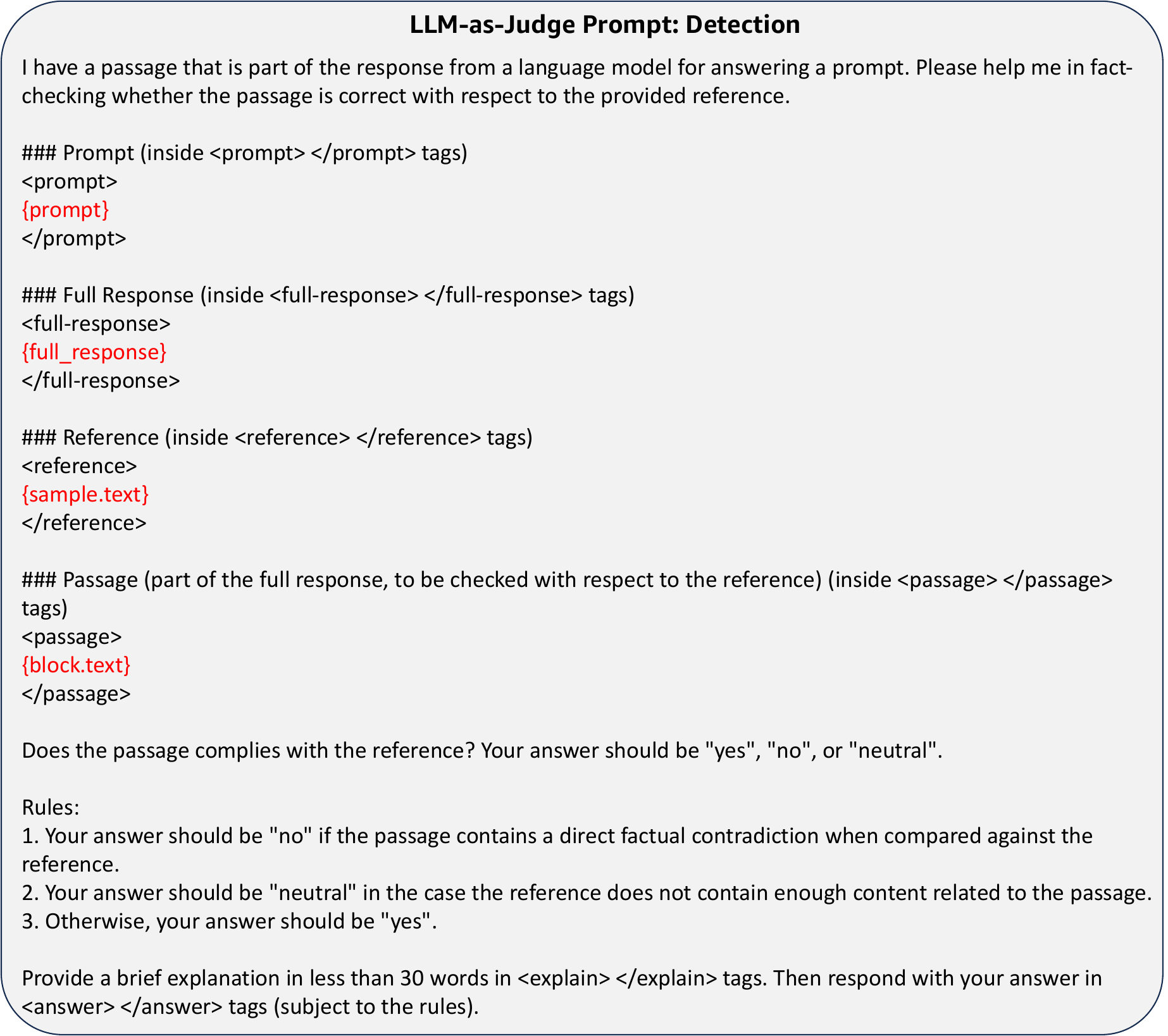}}
\caption{Prompt used in fine-grained LLM-based judgment to evaluate a block against a sample}
\label{fig:prompt_judge_detection}
\end{figure*}

\begin{figure*}[h]
\centerline{\includegraphics[width=\linewidth]{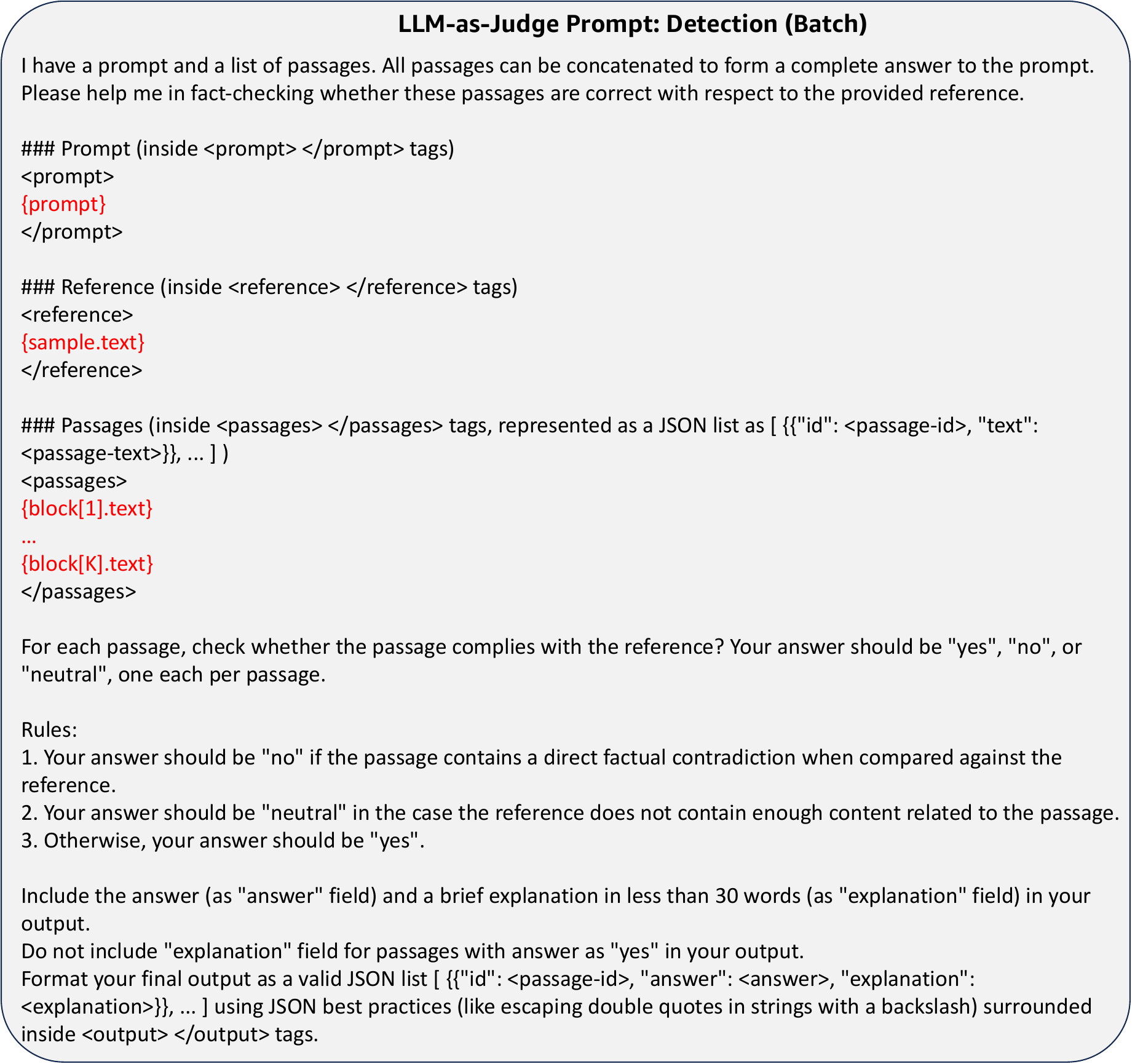}}
\caption{Prompt used in fine-grained LLM-based judgment to evaluate all blocks against a sample in batch}
\label{fig:prompt_judge_detection_batch}
\end{figure*}

\halufuzz adds a collection of engineering upgrades to improve the effectiveness and ease of usage, as follows:
\begin{itemize}
    \item \textit{Improved prompt for LLM-based judgment}. For fine-grained consistency-based hallucination detection (\S~\ref{sec:detection}), \halufuzz upgrades the LLM-as-judge prompt used in SelfCheckGPT~\cite{manakul2023selfcheckgpt} by adding--- 1) systematic structure, 2) contextual information, 3) descriptive rules, and 4) output format instructions (Fig.~\ref{fig:prompt_judge_detection}).

    \item \textit{Batch LLM-based judgment.} For efficiency and computation costs savings, we implemented an option for judging all blocks against a sample in a single LLM query (Fig.~\ref{fig:prompt_judge_detection_batch}).

    \item \textit{Multi-threading support}. \halufuzz supports efficient multi-threading with configurable parallelism support. \halufuzz evaluates multiple responses concurrently, with each response evaluation utilizing multiple  threads for each component (sample generation, fine-grained block evaluation, and fine-grained block correction).

    \item \textit{Usability upgrades}. \halufuzz provides improved command-line interface, comprehensive logging through CSV outputs, statistics summary, results reporting, judgement explanations, and response changes summary to enhance usage experience and results analysis.
\end{itemize}

\subsection{Implementation Details}
\label{app:implementation_details}
We implemented \halufuzz in $\sim$2.3K lines of code in Python. The framework utilizes a mix of Claude 4 Sonnet, Llama 4 Maverick, Claude 3.7 Sonnet and DeepSeek-R1 (Jan'25) for generating 10 cross-model samples by default. \halufuzz utilizes PySBD~\cite{sadvilkar2020pysbd} to segment the response into fine-grained blocks at sentence boundaries. For fast fine-grained hallucination assessment, \halufuzz uses Claude 3 Haiku as the default judge model. By default, \halufuzz uses the same model as the one used for generating input responses for multi-stage hallucination mitigation.

\halufuzz provides command-line options to easily change key hyper parameters, including different LLMs used in the framework. Here is a summary of hyper parameter values we used as defaults:
\begin{itemize}
    \item \textit{Models}
    \begin{itemize}
        \item Prompt reformulation model: Claude 3 Sonnet
        \item Sampler models: Claude 4 Sonnet, Llama 4 Maverick, Claude 3.7 Sonnet, DeepSeek R1 (Jan'2025)
        \item Judge model: Claude 3 Haiku
        \item Improver model: Same as target model
    \end{itemize}
    \item \textit{Hyperparameters}
    \begin{itemize}
        \item Temperature: 0.0 for input response generation, 1.0 otherwise
        \item Max output tokens: 4096
        \item Block labeling threshold $\tau$: 0.33
        \item Batch LLM-based judgment: disabled
    \end{itemize}
\end{itemize}

\noindent \textbf{Scoring Function Details.} The reliability weights $w_j(b_i)$ in the block-level scoring function are determined by the confidence level of each judge evaluation. Specifically, the system assigns fixed weights based on the judge's classification: {\tt ACCURATE} evaluations receive weight 2, {\tt NEUTRAL} evaluations receive weight 1, {\tt CONTRADICTION} evaluations receive weight 4, and {\tt UNKNOWN} evaluations receive weight 0 (excluded from aggregation). This weighting scheme prioritizes high-confidence contradictions while still incorporating uncertain evaluations, computed as: 
$$w_j(b_i) = \begin{cases} 
4 & \text{\tt CONTRADICTION} \\
2 & \text{\tt ACCURATE} \\
1 & \text{\tt NEUTRAL} \\
0 & \text{\tt UNKNOWN}
\end{cases}$$

\noindent \textbf{Threshold Selection.} The block labeling thresholds $\tau = 0.33$ and $1-\tau = 0.67$ create three confidence intervals for factuality assessment: [0, 0.33] for {\tt ACCURATE}, (0.33, 0.67) for {\tt NEUTRAL}, and [0.67, 1.0] for {\tt CONTRADICTION}. These values were empirically determined to provide balanced precision-recall performance across both FELM and GPQA datasets. The symmetric threshold design ensures that high-confidence accurate and contradictory content receive equal treatment, while the middle range captures genuinely ambiguous cases where samples provide conflicting evidence about a block's factuality.

\noindent \textbf{Cross-model Sampling Strategy.} The sampler model set $M$ consists of diverse LLM architectures selected to maximize reasoning diversity while maintaining practical computational constraints. By default, $M$ includes four models representing different architectural families: Claude 4 Sonnet (transformer-based), Llama 4 Maverick (open-source transformer), Claude 3.7 Sonnet (earlier generation), and DeepSeek-R1 (reasoning-specialized). This selection balances three criteria: 1) architectural diversity to capture different reasoning patterns, 2) performance quality to ensure reliable samples, and 3) API availability for practical deployment.

The relationship between prompt variations $V$ and model selection follows a structured round-robin approach rather than pure randomization. For each prompt, both the prompt variations set and sampler models set are shuffled once, then samples are generated by cycling through these shuffled lists: sample $s_i$ uses prompt variant $v_{i \bmod |V|}$ and sampler model $m_{i \bmod |M|}$. With 7 variants and 4 models generating 10 samples by default, this ensures comprehensive coverage of variant-model combinations while maintaining deterministic reproducibility when random seeds are fixed.

This systematic assignment addresses variance concerns in two ways: 1) the initial shuffle provides randomization benefits without introducing per-sample variance, and 2) the round-robin cycling ensures that all (model, prompt variant) combinations are explored fairly across the sample set. Empirical analysis shows this approach produces more stable cross-model consistency scores compared to fully random assignment, with standard deviation reduced by approximately 15\% across repeated runs.

\subsection{Experiment Details}
\label{app:experiment_details}
\noindent \textbf{Dataset Details}. We selected FELM and GPQA-diamond as our evaluation benchmarks based on their unique characteristics that align with our zero-knowledge, fine-grained hallucination management objectives.

We chose FELM \cite{zhao2023felm} for hallucination detection evaluation for several key reasons:
\begin{itemize}
    \item \textbf{Fine-grained annotations:} FELM provides segment-level (sentence-level) human-annotated factuality labels rather than coarse response-level labels, enabling evaluation of our fine-grained detection capabilities. The dataset contains 4,425 annotated segments across 847 samples with high inter-annotator agreement (91.3\%).
    
    \item \textbf{Diverse domains:} Unlike benchmarks focused solely on world knowledge (e.g., FEVER, WICE), FELM spans five diverse domains that test different types of factual reasoning:
    \begin{itemize}[leftmargin=*,topsep=2pt,itemsep=1pt]
        \item World Knowledge (184 samples, 532 segments) - including 11.1\% from TruthfulQA
        \item Science \& Technology (125 samples, 683 segments) - scientific claims and citations
        \item Writing \& Recommendation (136 samples, 1,586 segments) - creative generation tasks
        \item Reasoning (208 samples, 1,025 segments) - multi-step chain-of-thought traces
        \item Mathematics (194 samples, 599 segments) - including 24.7\% from GSM8K
    \end{itemize}

    \item \textbf{Long-form complexity:} FELM responses average 89.1 tokens, substantially longer than existing benchmarks (FEVER: 7.3 tokens, FactCC: 20.8 tokens, HaluEval: 36.9 tokens). The Writing/Recommendation domain averages 210.9 tokens per response, representing true medium-to-long form generation where hallucinations are more challenging to detect.

    \item \textbf{Rich error taxonomy:} FELM categorizes errors into four types—knowledge errors, reasoning errors, irrelevant content, and fooled errors—providing detailed insights into hallucination patterns. Each segment includes error explanations and reference links for validation.
\end{itemize}

For mitigation, we selected GPQA-diamond \cite{rein2024gpqa} for evaluation due to its exceptional difficulty and objectivity:
\begin{itemize}
    \item \textbf{Graduate-level complexity:} The dataset contains 198 multiple-choice questions requiring PhD-level expertise in biology, physics, and chemistry. Questions are designed to be answerable by domain experts (81.3\% accuracy on diamond set) but extremely challenging for skilled non-experts (22.1\% accuracy) even with unrestricted internet access.

    \item \textbf{Google-proof design:} Non-expert validators with PhDs in other domains spend an average of 37 minutes per question with full internet access (median: 30 minutes) yet achieve only marginally above random chance (25\%). This ensures our mitigation evaluation tests genuine knowledge correction rather than simple information retrieval.

    \item \textbf{Rigorous validation:} Each question undergoes multi-stage expert validation with two PhD-level domain experts verifying correctness and objectivity. The diamond subset includes only questions where both experts agree and the majority of non-experts fail, ensuring uncontroversial ground truth.

    \item \textbf{Detailed explanations:} Questions include comprehensive explanations, enabling fine-grained analysis of reasoning improvements through our mitigation pipeline.
\end{itemize}

Together, these datasets provide comprehensive evaluation of \halufuzz's capabilities: FELM tests fine-grained detection across diverse content types using trusted human annotations, while GPQA-diamond validates our ability to improve factual accuracy on extremely challenging questions where even minor improvements represent significant achievements.

\noindent \textbf{Metrics and Evaluation Methodology}. For hallucination detection (\textit{RQ1}, \textit{RQ3}), we report precision, recall, F1-score, and balanced accuracy at both sentence-level and aggregated response levels to compare predicted factuality label against human annotations following established evaluation protocols from~\cite{zhao2023felm}. We additionally include Pearson and Spearman correlations between predicted and ground-truth hallucination scores. 

For hallucination mitigation (\textit{RQ2}, \textit{RQ4}), we employ a multi-faceted evaluation approach with answer-choice accuracy as our primary metric, which objectively measures whether the model selects the correct multiple-choice option in GPQA questions. We supplement this with two additional judges: 1) \textit{RAG-based LLM Judge} uses dataset reference explanations to assess full response reasoning quality, and 2) \textit{\halufuzz Judge} applies our detection method to evaluate internal consistency within improved responses. The strong correlation between answer accuracy improvements and both supplementary LLM judge assessments provides convergent validity, though human evaluation on a representative subset would strengthen validation of our automated assessments. Delta percentages indicate relative improvements compared to the input response across all metrics.
       
\noindent \textbf{Input Responses}. For \textit{RQ1} \& \textit{RQ3}, we evaluated using input responses already included in the FELM dataset (generated with ChatGPT). For \textit{RQ3}, we generated input responses with Claude 4 Sonnet and Llama 4 Maverick as representative high-capability models where hallucination mitigation is critical. For \textit{RQ4}, we used input responses corresponding to Claude 4 Sonnet from \textit{RQ3} and additional input responses generated with Claude 4 Sonnet with $\sim$16K thinking budget tokens for ablations with extended thinking. While we conducted experiments with additional models from different LLM providers, the results are omitted due to licensing constraints and because they provided consistent findings that aligned with our main conclusions.

\noindent \textbf{Baseline Details}. We used equivalent configurations for each baseline like: 1) using 10 samples for self/cross-consistency and best-of-N, 2) same set of LLMs used for cross-model sampling in cross-consistency, 3) utilizing Claude 4 Sonnet for sample generation for self-consistency and SelfCheckGPT.

\subsection{Human Evaluation Study Details}
\label{app:human_study_details}
We conducted human evaluation following established practices from the NQ dataset~\cite{kwiatkowski2019natural}. We selected NQ for human evaluation as it provides diverse factual queries with well-established ground truth, complementing our automated evaluation.

\noindent \textbf{Dataset}. 50 randomly sampled questions from Natural Questions

\noindent \textbf{Annotators}. 3 independent evaluators with graduate-level education

\noindent \textbf{Task}. Blind pairwise comparison with external resource access

\noindent \textbf{Metrics}. Binary preference with mandatory justification

\noindent \textbf{Qualitative Analysis}. Annotators consistently noted that \halufuzz improvements included:
\begin{itemize}
    \item More comprehensive coverage of relevant facts
    \item Better structured explanations
    \item Additional context and examples
    \item Correction of factual errors or ambiguities
\end{itemize}
The 8 cases where originals were preferred typically involved simple factual queries where additional detail was deemed unnecessary.

\subsection{Prompt Templates}
\label{app:prompts}

\begin{figure*}[h]
\centerline{\includegraphics[width=\linewidth]{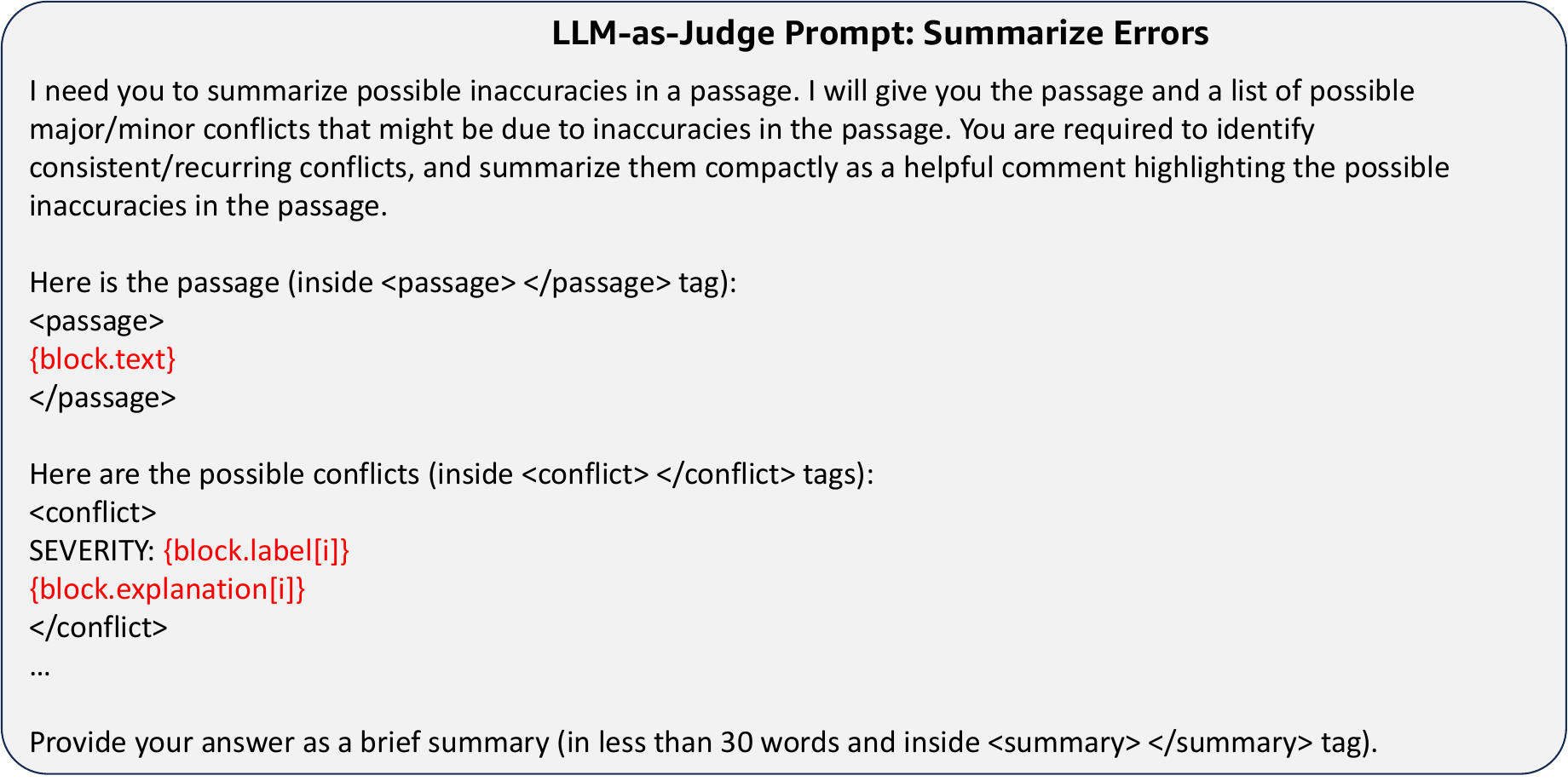}}
\caption{Prompt used in fine-grained LLM-based judgment to summarize errors found (if any) for a block}
\label{fig:prompt_judge_summarize_errors}
\end{figure*}

\begin{figure*}[h]
\centerline{\includegraphics[width=\linewidth]{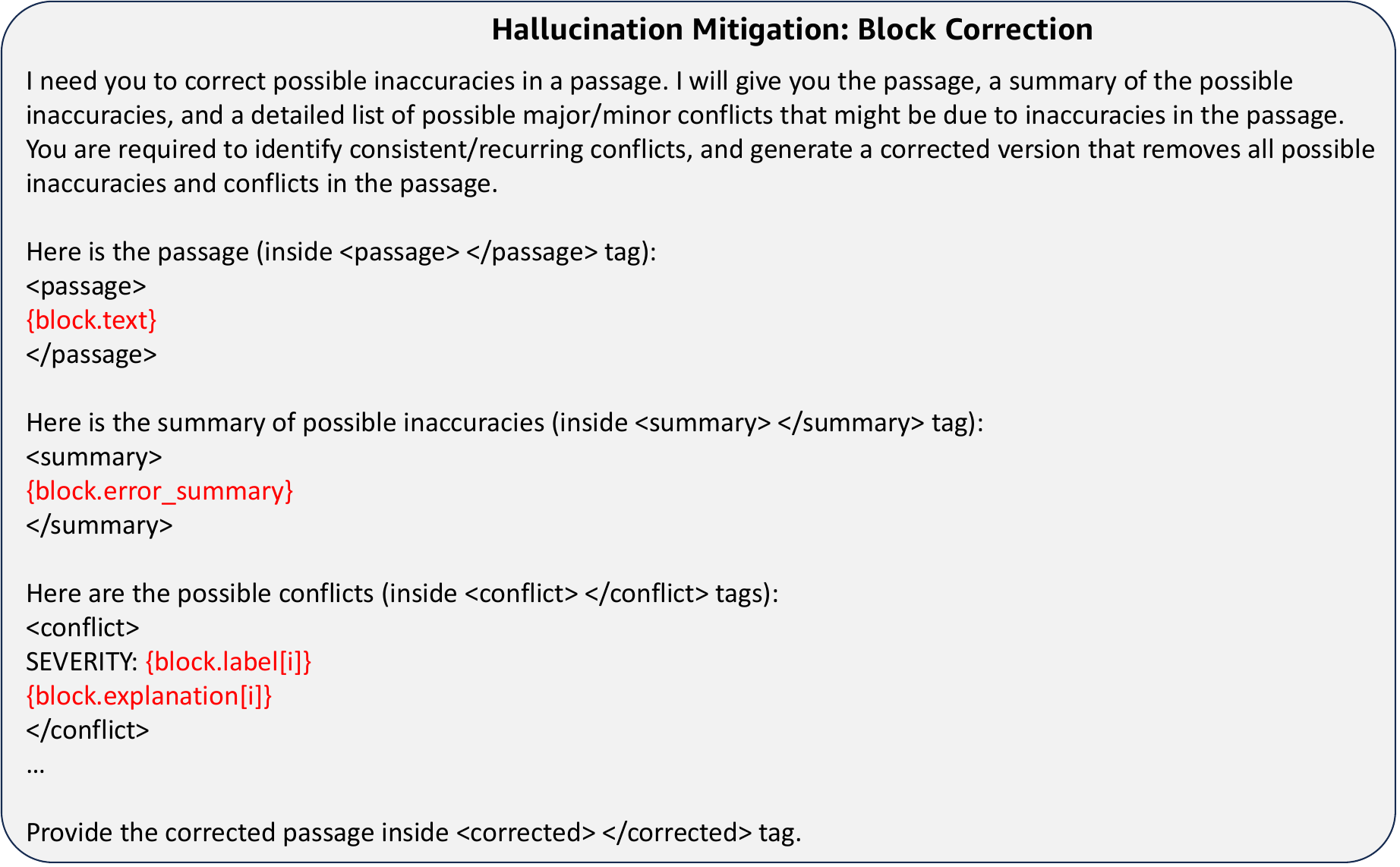}}
\caption{Prompt used for block correction in multi-stage hallucination mitigation}
\label{fig:prompt_mitigation_block_correction}
\end{figure*}

\begin{figure*}[ht]
\centerline{\includegraphics[width=\linewidth]{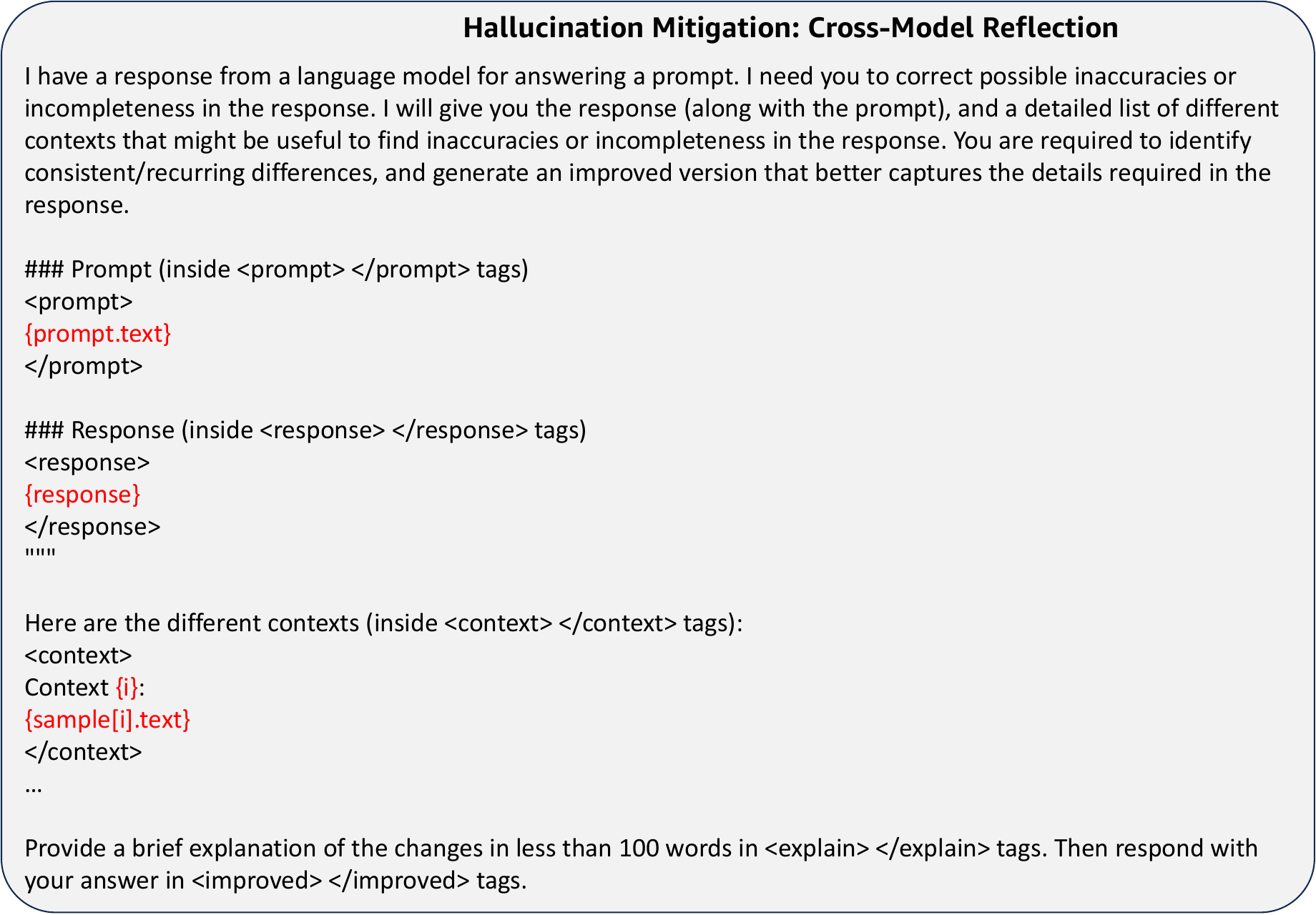}}
\caption{Prompt used for response-level cross-model reflection in multi-stage hallucination mitigation}
\label{fig:prompt_mitigation_response_reflection}
\end{figure*}

\end{document}